\documentclass[lettersize,journal]{IEEEtran}
\usepackage{amsmath,amsfonts,amssymb}
\usepackage{algorithmic}
\usepackage{algorithm}
\usepackage{array}
\usepackage{booktabs}
\usepackage{multirow}
\usepackage{makecell}
\usepackage{textcomp}
\usepackage{stfloats}
\usepackage{url}
\usepackage{graphicx}
\usepackage{cite}
\usepackage{xcolor}
\usepackage{tikz}
\usetikzlibrary{arrows.meta,positioning,fit,calc}
\usepackage[hidelinks]{hyperref}
\newcommand{\src}{\mathrm{s}}
\newcommand{\BN}{\mathrm{BN}}
\newcommand{\Enc}{E_{\theta}}
\newcommand{\Fuse}{\Phi}
\newcommand{\Head}{H}
\newcommand{\bmu}{\boldsymbol{\mu}}
\newcommand{\bsig}{\boldsymbol{\sigma}}
\newcommand{\ours}{AgentNorm}
\newcommand{\apf}{AP@0.5}
\newcommand{\aps}{AP@0.7}

\begin{document}

\title{Agent-Wise Test-Time Normalization for Camera-Based Cooperative Perception with Corrupted Collaborators}

\author{Yu~Wang, Shaohua~Wang, Yicheng~Li, and Kecheng~Sun
\thanks{This work has been submitted to the IEEE for possible publication. Copyright may be transferred without notice, after which this version may no longer be accessible.}%
\thanks{Yu Wang, Shaohua Wang, Yicheng Li, and Kecheng Sun are with the Institute of Automotive Research, Jiangsu University, Zhenjiang 212013, China (e-mail: wangyu@stmail.ujs.edu.cn; wsh@ujs.edu.cn; liyicheng070@ujs.edu.cn; sunkc@ujs.edu.cn).}%
\thanks{Corresponding author: Shaohua Wang (e-mail: wsh@ujs.edu.cn).}}

\markboth{Preprint submitted to IEEE Transactions on Vehicular Technology}%
{Wang \MakeLowercase{\textit{et al.}}: Agent-Wise Test-Time Normalization for Cooperative Perception}

\maketitle

\begin{abstract}
Intermediate-fusion cooperative perception lets a vehicle detect objects beyond its own field of view by fusing features shared by nearby agents. When only a collaborator's cameras are degraded by fog, motion blur or exposure errors, the features it shares drift away from the training distribution while the ego input stays clean, and cooperation can become worse than not cooperating at all. Test-time batch-normalization adaptation, the standard training-free remedy, estimates statistics over the pooled batch of all agents and therefore normalizes both the clean ego and the corrupted collaborator with a mixture of two distributions. We show that this mixing harms even the clean ego: when the ego shares BN statistics with a corrupted collaborator, its detection from its own features drops by 3.58 to 6.68 \apf{} points although its cameras are untouched. We propose \ours{}, an agent-wise test-time normalization that i) estimates the normalization statistics of every collaborator from its own current observation and ii) anchors the ego to its source statistics, so that the ego path remains exactly the original model. \ours{} has no learnable parameters, needs no gradients or labels, is independent of the order of test frames, and runs locally on each agent without changing the transmitted features. On camera-based OPV2V with four fusion checkpoints and collaborator-only corruptions, \ours{} raises \apf{} under severity-5 fog from 29.42--37.25 to 43.32--58.86 and outperforms pooled normalization by 5.53--12.94 points under fog and motion blur, at an extra latency of 2.3~ms per frame. On the real-world DAIR-V2X-C dataset, it outperforms pooled normalization in all ten infrastructure-corruption settings. We also report the cases it does not fix: occlusion, which removes image content rather than shifting feature statistics, and the case in which all agents share the same shift.

\end{abstract}


\section{Introduction}
\label{sec:intro}

\IEEEPARstart{C}{ooperative} perception allows connected vehicles and roadside units to share what they observe, so that an ego vehicle can detect objects that are occluded or out of range of its own sensors~\cite{xu2021opv2v,xu2022v2xvit,hu2022where2comm}. Among the fusion strategies, intermediate fusion, in which every agent encodes its own sensor data into a bird's-eye-view (BEV) feature map and transmits it to the ego, gives the best trade-off between accuracy and bandwidth and is the default in recent frameworks~\cite{lu2024heal}. Camera-only cooperative perception is attractive for large-scale deployment because cameras are cheap and already installed on most vehicles and many intersections. Cameras, however, are also the sensor most exposed to adverse weather, dirt, glare and vibration~\cite{du2025rcpbench,xie2023robobev}. In a cooperative system these degradations rarely hit all agents at once: a roadside camera may face the sun while the ego does not, or a collaborator may drive through a fog bank that the ego has already left. The distribution shift is \emph{agent-specific}.

Agent-specific corruption breaks an assumption that most cooperative perception methods make implicitly: that every received feature map comes from the training distribution. When a collaborator's cameras are corrupted, its BEV feature drifts away from what the fusion module expects, and the fused prediction degrades. On the camera-based OPV2V benchmark, an attention-fusion detector receiving a collaborator under severity-5 fog reaches 37.25 \apf{}, which is below the 38.31 obtained by simply discarding the collaborator (Section~\ref{sec:analysis}). RCP-Bench~\cite{du2025rcpbench} documents large accuracy drops of camera-based cooperative detectors under a wide range of image corruptions.

Several remedies are available, but each has a cost in this setting. Training with corruption augmentation~\cite{du2025rcpbench} or with domain-generalization objectives~\cite{li2025v2xdgw,zhang2025dsrc} requires retraining every deployed model and covering every corruption in advance. Discarding a collaborator that looks unreliable gives up the benefit of cooperation. Gradient-based test-time adaptation (TTA)~\cite{wang2020tent,wang2022cotta} updates parameters online without labels, but a Tent-style entropy objective on the dense detection map of a camera BEV detector collapsed in our runs: \apf{} under collaborator fog falls from 37.25 to 3.94 or lower, and the collapse propagates to the ego through fusion even when only the collaborator's parameters are updated (Section~\ref{sec:analysis}). The remaining training-free option is to re-estimate batch-normalization (BN) statistics on the test data~\cite{bnadapt,mirza2022dua}, which RCP-Bench uses as its test-time baseline~\cite{du2025rcpbench}. Applied to a cooperative model, this computes BN statistics over the pooled batch formed by all agents of a frame. Under agent-specific corruption, the pooled statistics are a mixture of a clean and a corrupted distribution and match neither. We find that this mixing contaminates the clean ego: detecting from the ego's own features alone, its \apf{} drops from 38.31 to between 31.63 and 34.73 once it shares statistics with a corrupted collaborator (Section~\ref{sec:pooled}).

Our key observation is that the collaborator's corruption is visible in its own feature statistics. As the severity increases, the deviation of the collaborator's BEV features from the source statistics grows monotonically and their local variance shrinks, for all four corruption types we tested (Fig.~\ref{fig:stats}). The ego's features are unaffected. The statistics needed to correct the collaborator are therefore available in the collaborator's own observation, and they should not be mixed with the ego's. Based on this observation, we propose \ours{}, an agent-wise test-time normalization with two components. \emph{Agent-wise statistics} split every BN input by agent and normalize each collaborator with the mean and variance of its own current observation. \emph{Ego anchoring} keeps the ego on its source statistics, so that the ego path is the unmodified pre-trained model. \ours{} adds no parameters, takes no gradient steps, uses no labels, and gives the same result regardless of the order of test frames. Because each agent normalizes its own features before transmission, it also changes neither the message format nor the bandwidth.

We evaluate \ours{} on camera-based OPV2V~\cite{xu2021opv2v} with four public fusion checkpoints and on the real-world DAIR-V2X-C dataset~\cite{yu2022dairv2x}. Under severity-5 fog and motion blur of the collaborator, \ours{} improves \apf{} over pooled normalization by 5.53 to 12.94 points on every checkpoint, and it recovers 74\% of the gap between the unadapted model and an oracle that adapts the collaborator with ground-truth supervision. The extra latency is 2.3~ms per frame on top of 254~ms.

The contributions of this paper are as follows.
\begin{itemize}
\item We identify agent-specific corruption as a distinct test-time setting for cooperative perception and show that pooled test-time BN, the standard training-free baseline, contaminates the clean ego with the statistics of a corrupted collaborator. We quantify this contamination and locate it in the image encoder. We also show that region-level gating and entropy-based TTA fail in this setting.
\item We propose \ours{}, which combines agent-wise statistics with ego anchoring. It is training-free, gradient-free and communication-free, and it can be dropped into any intermediate-fusion model whose per-agent encoder uses BN.
\item We conduct experiments on four fusion checkpoints, four corruption types at two severities, and a real-world dataset, and we report the conditions under which \ours{} does not help, including occlusion and agent-homogeneous shift.
\end{itemize}

\section{Related Work}
\label{sec:related}

\subsection{Cooperative Perception}
Cooperative perception datasets such as OPV2V~\cite{xu2021opv2v}, DAIR-V2X~\cite{yu2022dairv2x} and V2V4Real~\cite{xu2023v2v4real} have made intermediate fusion the dominant paradigm. F-Cooper~\cite{chen2019fcooper} fuses voxel features by max pooling, AttFuse~\cite{xu2021opv2v} applies per-location self-attention across agents, V2X-ViT~\cite{xu2022v2xvit} models heterogeneous vehicle and infrastructure agents with a vision transformer, DiscoNet~\cite{li2021disconet} distills a collaboration graph from an early-fusion teacher, and Where2comm~\cite{hu2022where2comm} reduces bandwidth by transmitting only spatially confident regions. HEAL~\cite{lu2024heal} provides an extensible framework and public camera and LiDAR checkpoints for these fusion methods, which we use in our experiments. Later work has addressed practical non-idealities, including communication latency~\cite{lei2022syncnet}, lossy channels~\cite{li2022v2vam}, and the domain gap between agents running different models~\cite{xu2023mpda}. These methods assume that the sensor input of every agent is drawn from the training distribution. We study the case in which this assumption fails for some agents only, and we leave the fusion architecture unchanged.

\subsection{Robustness of Cooperative Perception to Corruptions}
Corruption benchmarks for single-agent 3D detection~\cite{hendrycks2019benchmarking,dong2023corrupt3d,xie2023robobev} have been extended to the cooperative setting by RCP-Bench~\cite{du2025rcpbench}, which evaluates camera-based cooperative models under diverse image corruptions and proposes training strategies to improve robustness. Adver-City~\cite{karvat2024advercity} provides cooperative data under adverse weather. Training-time remedies include domain generalization under adverse weather~\cite{li2025v2xdgw}, density-insensitive and semantic-aware representations~\cite{zhang2025dsrc}, and diffusion-guided feature restoration~\cite{chen2026coopdiff}. Sim-to-real adaptation of cooperative perception has been studied with unsupervised domain adaptation~\cite{li2024s2rvit,kong2023dusa} and with test-time adaptation~\cite{wang2026v2xtta}, and CoPEFT~\cite{wei2025copeft} adapts cooperative models with parameter-efficient fine-tuning. All of these either change the training procedure or treat the domain shift as shared by all agents. \ours{} requires no retraining and is designed for shifts that differ between agents.

\subsection{Test-Time Adaptation and Test-Time Normalization}
Fully test-time adaptation updates a source model on unlabeled test data~\cite{liang2023ttasurvey,wang2023ottasurvey}. Tent~\cite{wang2020tent} minimizes prediction entropy with respect to the BN affine parameters, EATA~\cite{niu2022eata} filters unreliable samples to avoid forgetting, CoTTA~\cite{wang2022cotta} and RoTTA~\cite{yuan2023rotta} target continually changing test streams, and test-time training adds a self-supervised auxiliary task~\cite{sun2019ttt}. For 3D detection, MonoTTA~\cite{lin2024monotta} and DUO~\cite{hu2025duo} adapt monocular detectors, and V2X-TTA~\cite{wang2026v2xtta} adapts cooperative detectors from simulation to real data. A lighter family only re-estimates normalization statistics: replacing the BN statistics with those of the test batch already removes much of the error under common corruptions~\cite{bnadapt}, DUA~\cite{mirza2022dua} updates them with a running average, ActMAD~\cite{mirza2023actmad} aligns activation statistics with those of the training data, and domain-aware BN~\cite{dabn} learns how to combine the two. Test-time BN has also been applied to object detection for driving~\cite{liao2024bntta}. All these methods treat the test batch as a sample from a single target distribution. In a cooperative model the batch of a BN layer contains features of several agents whose distributions can differ, and \ours{} differs from prior test-time normalization in computing the statistics per agent and keeping the ego at its source statistics.

Table~\ref{tab:setting} summarizes how our setting relates to existing adaptation settings. It keeps the constraints of test-time BN (no source data, no labels, no retraining, no gradients) and differs in one column only: the shift may affect a subset of the agents that are processed together.

\begin{table}[t]
\centering
\caption{Adaptation settings. Our setting keeps the constraints of test-time BN and adds that the shift may affect only a subset of the agents processed by the same network.}
\label{tab:setting}
\setlength{\tabcolsep}{2.6pt}
\footnotesize
\begin{tabular}{lccccc}
\toprule
Setting & \makecell{Source\\data} & \makecell{Target\\labels} & \makecell{Re-\\training} & \makecell{Test-time\\gradients} & \makecell{Agent-\\specific shift} \\
\midrule
Fine-tuning & -- & \checkmark & \checkmark & -- & -- \\
Domain adaptation & \checkmark & -- & \checkmark & -- & -- \\
Domain generalization & \checkmark & -- & \checkmark & -- & -- \\
Test-time training & \checkmark & -- & \checkmark & \checkmark & -- \\
Fully TTA (e.g., Tent) & -- & -- & -- & \checkmark & -- \\
Test-time BN & -- & -- & -- & -- & -- \\
\textbf{Ours} & -- & -- & -- & -- & \checkmark \\
\bottomrule
\end{tabular}
\end{table}

\section{Preliminaries and Problem Setting}
\label{sec:problem}

\subsection{Camera-Based Intermediate Fusion}
We consider a frame with $N$ agents indexed by $i\in\{0,1,\dots,N-1\}$, where $i=0$ denotes the ego. Agent $i$ captures $C$ images $\mathcal{I}_i=\{I_i^{c}\}_{c=1}^{C}$. A per-agent encoder $\Enc$, shared by all agents, lifts the images into a BEV feature map
\begin{equation}
F_i=\Enc(\mathcal{I}_i)\in\mathbb{R}^{D\times H\times W}.
\end{equation}
In the camera models studied here, $\Enc$ consists of an image backbone, a Lift-Splat-Shoot view transformer~\cite{philion2020lss} and a BEV backbone. Collaborators transmit $F_i$ to the ego, where it is warped into the ego coordinate frame by $\mathcal{T}_{i\to0}$. A fusion module $\Fuse$ and a detection head $\Head$ produce the prediction
\begin{equation}
\hat{Y}=\Head\big(\Fuse(F_0,\{\mathcal{T}_{i\to0}(F_i)\}_{i=1}^{N-1})\big).
\end{equation}

\subsection{Batch Normalization in the Per-Agent Encoder}
Each BN layer of $\Enc$ receives a tensor $X\in\mathbb{R}^{B\times K\times h\times w}$ with $K$ channels. During training, a BN layer normalizes $X$ with the statistics of the current mini-batch and accumulates running estimates $(\bmu_\src,\bsig^2_\src)$. At inference, the source statistics are used:
\begin{equation}
\BN(x;\bmu,\bsig^2)=\gamma\,\frac{x-\bmu}{\sqrt{\bsig^2+\epsilon}}+\beta,\qquad (\bmu,\bsig^2)=(\bmu_\src,\bsig^2_\src),
\label{eq:bn}
\end{equation}
where $\gamma$ and $\beta$ are learned affine parameters and all operations are channel-wise. In a cooperative model, the encoder processes the agents of a frame together, so the batch dimension $B$ contains the slices of several agents: the $C$ images of each agent in the image backbone and one BEV map per agent in the BEV backbone. We write $X=[X_0;X_1;\dots;X_{N-1}]$ for this agent-major layout. In the AttFuse checkpoint used in Section~\ref{sec:analysis}, all 75 BN layers lie in $\Enc$; the fusion module and the detection head contain no normalization layer.

\subsection{Agent-Specific Corruption at Test Time}
The model is trained on clean data from a source distribution $P_\src$. At test time, frames arrive one by one. The ego's images follow $P_\src$, whereas the images of collaborator $i$ follow a corrupted distribution $Q_i$ that may differ from $P_\src$ and from the other collaborators. The ego must output detections for each frame under the following constraints: i) no labels and no source data are available; ii) the model is not retrained; iii) every frame is processed once, with a latency close to that of the original model; and iv) the message transmitted between agents keeps its format and size. We additionally prefer methods without gradient updates, since we show in Section~\ref{sec:analysis} that gradient-based adaptation is unstable for the detectors considered. The goal is to maximize the ego's detection accuracy.

\section{Empirical Analysis}
\label{sec:analysis}

Before introducing the method, we examine what goes wrong when only a collaborator is corrupted and why the obvious remedies do not fix it. All experiments in this section use the public camera AttFuse checkpoint of HEAL~\cite{lu2024heal} on the OPV2V test split; the setup is detailed in Section~\ref{sec:setup}.

\subsection{A Corrupted Collaborator Can Be Worse Than None}
Table~\ref{tab:gating} lists the \apf{} of cooperative detection when only the collaborators' cameras are corrupted (column ``No adapt.'') and when the collaborators are discarded (``Ego only''). With clean collaborators, the model reaches 56.81 \apf{}. Under severity-5 fog and motion blur, cooperation drops to 37.25 and 36.48, below the 38.31 of the ego alone. For the other six settings cooperation still helps, but the gain over the ego alone shrinks from 18.50 points with clean collaborators to between 0.54 and 14.69 points. The attention fusion therefore absorbs part of the damage but not all of it.

\subsection{Corruption Shifts the Collaborator's Feature Statistics}
\begin{figure}[t]
\centering
\includegraphics[width=\linewidth]{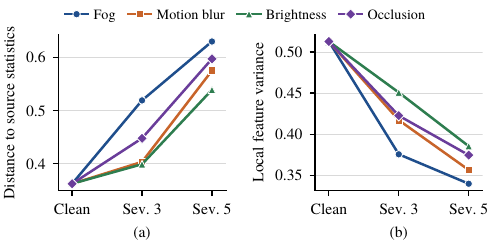}
\caption{Statistics of the collaborator's BEV features versus the severity of its camera corruption (AttFuse, OPV2V test, ego clean). (a) Mean normalized distance to the source statistics. (b) Mean local variance. Both change monotonically for all corruption types.}
\label{fig:stats}
\end{figure}

We measure, on every BEV cell covered by a collaborator's cameras, two statistics of the collaborator's pre-fusion feature $F_i$: the normalized distance to per-cell source statistics computed on clean training frames, and the local variance in a $3\times3$ neighborhood across all channels. Fig.~\ref{fig:stats} shows their means over the 3,445 collaborator instances of the test split. For all four corruptions, the distance to the source statistics increases monotonically with the severity (from 0.36 on clean data to between 0.54 and 0.63 at severity 5), and the local variance decreases monotonically (from 0.51 to between 0.34 and 0.39). The corruption thus appears as a systematic change of the first- and second-order statistics of the collaborator's features, which is precisely what normalization layers control.

\subsection{Why Not Gate Corrupted Regions?}
Since the shift is visible per BEV cell, a natural idea is to suppress the cells of the collaborator that deviate most from the source statistics before fusion. We score each covered cell by the distance above and set the top-$k\%$ cells to zero, which is a value the fusion module has seen during training for cells outside the cameras' view. Table~\ref{tab:gating} reports the result. The score does locate the cells whose features changed most: its AUROC against a reference label obtained from the paired clean frame is between 0.65 and 0.84. Yet gating by this score is worse than gating random cells of the same number in all 24 combinations of setting and $k$, and in only three of them, all under fog, is it better than both no gating and discarding the collaborator, by at most 0.24 points. Even oracle gating, which removes the cells that changed most according to the clean reference, improves over the better of no gating and ego-only by at most 2.31 points. Removing features cannot restore information, and cells whose statistics shifted most are not necessarily harmful to detection. This motivates repairing the collaborator's features as a whole rather than masking them locally.

\begin{table}[t]
\centering
\caption{Region-level gating of the collaborator (AttFuse, OPV2V, \apf{}). Only the collaborators' cameras are corrupted. AUROC: localization of shifted cells by the gating score. Gating columns report the best of $k\in\{10,30,50\}\%$. Underlined: cooperation below ego-only. Oracle uses the paired clean frame and is not a feasible method.}
\label{tab:gating}
\setlength{\tabcolsep}{2.8pt}
\footnotesize
\begin{tabular}{llcccccc}
\toprule
Corruption & Sev. & AUROC & \makecell{No\\adapt.} & \makecell{Ego\\only} & \makecell{Score\\gating} & \makecell{Random\\gating} & \makecell{Oracle\\gating} \\
\midrule
Fog & 3 & 0.719 & 38.85 & 38.31 & 39.09 & 39.50 & 41.16 \\
Fog & 5 & 0.743 & \underline{37.25} & 38.31 & 38.45 & 38.80 & 40.02 \\
Motion blur & 3 & 0.708 & 42.53 & 38.31 & 40.36 & 42.06 & 44.16 \\
Motion blur & 5 & 0.752 & \underline{36.48} & 38.31 & 38.14 & 38.25 & 39.91 \\
Brightness & 3 & 0.650 & 53.00 & 38.31 & 44.35 & 50.78 & 51.78 \\
Brightness & 5 & 0.719 & 42.37 & 38.31 & 41.01 & 41.69 & 42.28 \\
Occlusion & 3 & 0.777 & 46.99 & 38.31 & 42.42 & 45.83 & 47.56 \\
Occlusion & 5 & 0.835 & 39.92 & 38.31 & 39.16 & 39.65 & 40.36 \\
\bottomrule
\end{tabular}
\end{table}

\subsection{A Tent-Style Entropy Baseline}
We next adapt the BN affine parameters online with a Tent-style objective~\cite{wang2020tent}: the binary entropy of the anchors whose confidence exceeds a threshold is minimized, with one frame per step and Adam as the optimizer. Unlike the original Tent, the BN statistics stay at their source values, so only the affine parameters are adapted. We consider updating all BN layers and updating only a separate copy of the encoder used for the collaborators. Under severity-5 fog, \apf{} drops from 37.25 to 1.59 when all BN layers are updated and to 3.94 when only the collaborator's encoder is updated (learning rate $10^{-4}$, threshold 0.2); fog, motion blur and brightness behave the same, with \apf{} between 0.17 and 4.54 in all twelve runs at this threshold (two learning rates, $10^{-4}$ and $10^{-3}$). The collapse goes in either direction depending on the threshold. At threshold 0.2, entropy minimization pushes the many low-confidence anchors towards zero until no anchor exceeds the threshold. At threshold 0.5 (tested under fog), it pushes them towards one, and the all-BN model outputs on average 12,208 anchors per frame above 0.5, reaching 0.17 \apf{}. The run that adapts only the collaborator's encoder shows that a failure of the collaborator's adaptation is passed to the ego through fusion: the ego's own path is untouched, yet the fused output collapses. We do not claim that no gradient-based method can work here: we tested this objective with two thresholds, two learning rates and one seed, and did not test sample selection~\cite{niu2022eata} or periodic resets. The failure shows that a plain entropy objective on the dense binary map is unreliable in this setting, which motivates a method that is stable by construction.

\section{Agent-Wise Test-Time Normalization}
\label{sec:method}

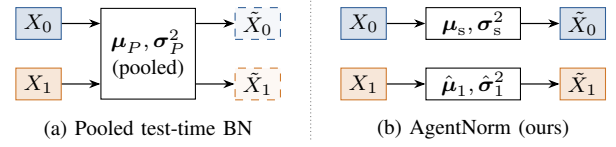
\begin{figure}[t]
\centering
\begin{tikzpicture}[
  font=\footnotesize,
  feat/.style={draw=#1, fill=#1!18, minimum width=0.6cm, minimum height=0.42cm, inner sep=0pt},
  op/.style={draw, minimum width=1.25cm, minimum height=0.42cm, inner sep=1pt, align=center},
  arr/.style={-{Stealth[length=1.5mm]}, line width=0.45pt},
]
\definecolor{egoc}{HTML}{2F5F98}\definecolor{colc}{HTML}{D0782A}
\node[feat=egoc] (e1) at (0,0.4) {$X_0$};
\node[feat=colc] (c1) at (0,-0.4) {$X_1$};
\node[op, minimum height=1.2cm] (p) at (1.45,0) {$\bmu_P,\bsig_P^2$\\[1pt](pooled)};
\node[feat=egoc, fill=egoc!5, dashed] (e1o) at (2.9,0.4) {$\tilde{X}_0$};
\node[feat=colc, fill=colc!5, dashed] (c1o) at (2.9,-0.4) {$\tilde{X}_1$};
\draw[arr] (e1.east) -- (e1.east -| p.west);
\draw[arr] (c1.east) -- (c1.east -| p.west);
\draw[arr] (e1o.west -| p.east) -- (e1o.west);
\draw[arr] (c1o.west -| p.east) -- (c1o.west);
\node at (1.45,-1.0) {(a) Pooled test-time BN};
\node[feat=egoc] (e2) at (4.3,0.4) {$X_0$};
\node[feat=colc] (c2) at (4.3,-0.4) {$X_1$};
\node[op] (s) at (5.75,0.4) {$\bmu_\src,\bsig_\src^2$};
\node[op] (o) at (5.75,-0.4) {$\hat\bmu_1,\hat\bsig_1^2$};
\node[feat=egoc] (e2o) at (7.2,0.4) {$\tilde{X}_0$};
\node[feat=colc] (c2o) at (7.2,-0.4) {$\tilde{X}_1$};
\draw[arr] (e2) -- (s); \draw[arr] (s) -- (e2o);
\draw[arr] (c2) -- (o); \draw[arr] (o) -- (c2o);
\node at (5.75,-1.0) {(b) \ours{} (ours)};
\draw[densely dotted, gray] (3.6,0.75) -- (3.6,-1.15);
\end{tikzpicture}
\caption{One BN layer of the per-agent encoder with a clean ego ($X_0$, blue) and a corrupted collaborator ($X_1$, orange). (a) Pooled test-time BN normalizes both agents with the mixture statistics $(\bmu_P,\bsig_P^2)$, which shifts the clean ego (dashed) and under-corrects the collaborator. (b) \ours{} keeps the ego on the source statistics (ego anchoring) and normalizes the collaborator with the statistics of its own current observation (agent-wise statistics).}
\label{fig:method}
\end{figure}

Section~\ref{sec:analysis} showed that collaborator corruption shifts the first- and second-order statistics of the collaborator's features, that removing features locally does not help, and that gradient-based adaptation is unstable. \ours{} corrects the statistics of each collaborator in closed form inside the existing BN layers. Fig.~\ref{fig:method} contrasts it with pooled test-time BN. We first show why pooling is the wrong estimator under agent-specific shift (Section~\ref{sec:pooled}) and then describe the two components of \ours{}: agent-wise statistics (Section~\ref{sec:agentwise}) and ego anchoring (Section~\ref{sec:anchor}).

\subsection{Pooled Test-Time Normalization Is Mis-Specified}
\label{sec:pooled}
Test-time BN~\cite{bnadapt} replaces $(\bmu_\src,\bsig_\src^2)$ in \eqref{eq:bn} by the statistics of the current test batch. In a cooperative model, the batch of a BN layer is $X=[X_0;\dots;X_{N-1}]$ when the encoder runs all agents of a frame together, as simulators and the RCP-Bench baseline do. Let $\bmu_i$ and $\bsig_i^2$ be the channel-wise mean and variance of the slice $X_i$, and $\pi_i$ its share of the elements of $X$. The pooled statistics are
\begin{align}
\bmu_P&=\sum_{i}\pi_i\bmu_i, \label{eq:pmu}\\
\bsig_P^2&=\sum_{i}\pi_i\bsig_i^2+\sum_{i}\pi_i(\bmu_i-\bmu_P)^2. \label{eq:pvar}
\end{align}
Normalizing agent $j$ with them gives
\begin{equation}
\frac{x-\bmu_P}{\bsig_P}=\frac{\bsig_j}{\bsig_P}\,z_j+\frac{\bmu_j-\bmu_P}{\bsig_P},\qquad z_j=\frac{x-\bmu_j}{\bsig_j},
\label{eq:residual}
\end{equation}
so every agent keeps a residual offset $(\bmu_j-\bmu_P)/\bsig_P$ and a scale error $\bsig_j/\bsig_P$ unless all agents share the same statistics. Note that i) the clean ego is affected as well: with one collaborator and equal shares, the ego is offset by $(\bmu_0-\bmu_1)/(2\bsig_P)$, half of the collaborator's deviation; ii) the between-agent term in \eqref{eq:pvar} inflates $\bsig_P$, so the collaborator's features are also shrunk; and iii) when all agents undergo the same shift, $\bmu_i=\bmu_P$ and the residuals vanish, so pooling becomes the right estimator. Point iii) predicts that the advantage of per-agent statistics should disappear when the shift is shared, which we verify in Section~\ref{sec:ablation}.

\begin{figure*}[t]
\centering
\includegraphics[width=0.86\textwidth]{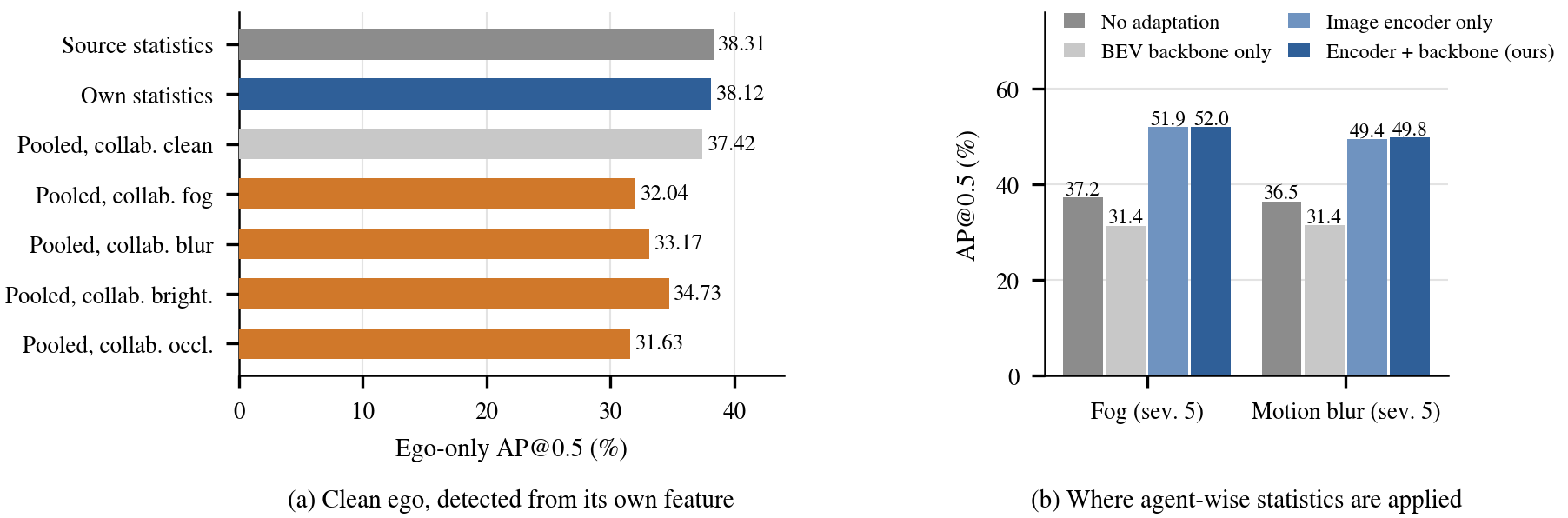}
\caption{Mechanism analysis on AttFuse (OPV2V, severity 5). (a) \apf{} of the clean ego detected from its own feature alone, under source statistics, its own statistics, and statistics pooled with a clean or corrupted collaborator. Pooling with a corrupted collaborator lowers the ego's own detection by 3.58--6.68 points. (b) \apf{} of cooperative detection when agent-wise statistics are applied only in the BEV backbone, only in the image encoder, or in both (\ours{}).}
\label{fig:mechanism}
\end{figure*}

\textbf{Pooling contaminates the clean ego.} Point i) can be measured directly. We run AttFuse with a clean ego and severity-5 corrupted collaborators, apply a given normalization, and then detect from the ego's own feature alone, discarding the collaborators before fusion. Any change in this ego-only accuracy can only come from the normalization of the ego's features. Fig.~\ref{fig:mechanism}(a) shows the result. With source statistics, the ego alone reaches 38.31 \apf{}, and with statistics computed from its own clean images it reaches 38.12. Once the ego shares pooled statistics with a corrupted collaborator, its own detection falls to 32.04 (fog), 33.17 (motion blur), 34.73 (brightness) and 31.63 (occlusion), that is, by 3.58 to 6.68 points, although its cameras are untouched. Pooling with a clean collaborator costs only 0.89 points. The corrupted collaborator therefore degrades the ego twice: through its own shifted features in fusion, and through the ego's normalization.

\begin{figure*}[t]
\centering
\includegraphics[width=0.92\textwidth]{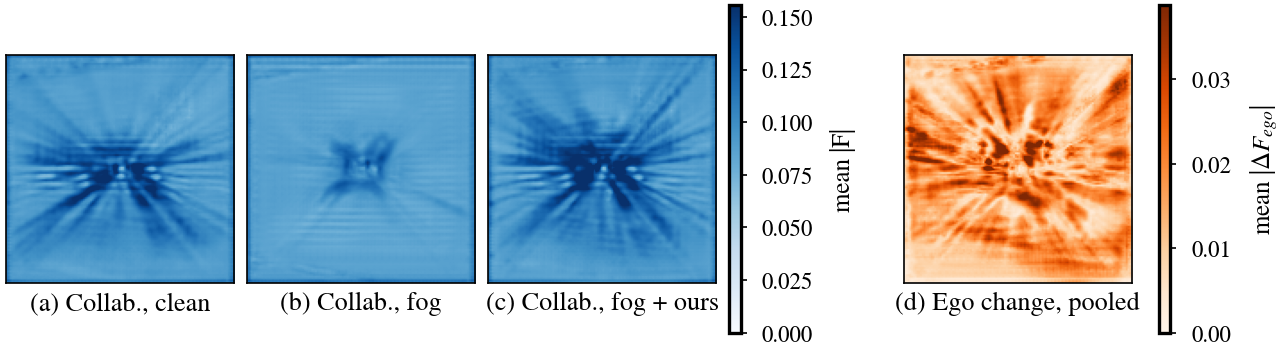}
\caption{BEV features on an OPV2V test frame (AttFuse) with the collaborator under severity-5 fog and a clean ego. (a)--(c) Channel-averaged magnitude of the collaborator's feature with clean cameras, with fog, and with fog after \ours{}. (d) Channel-averaged absolute change of the ego's feature relative to the source model under Pooled BN. Under \ours{} the same change averages $7.3\times10^{-5}$ (floating-point level, against 0.016 for Pooled BN) because the ego keeps its source statistics, so it is not plotted.}
\label{fig:features}
\end{figure*}

Fig.~\ref{fig:features} illustrates these effects on an OPV2V test frame whose collaborator is under severity-5 fog. Fog lowers the magnitude of the collaborator's BEV feature (Fig.~\ref{fig:features}(b); map average 0.090 versus 0.100 with clean input in (a)), consistent with the variance reduction in Fig.~\ref{fig:stats}(b). Agent-wise statistics restore the magnitude (0.104) and the spatial structure seen with the clean input (c). Pooled statistics, in contrast, change the features of the clean ego over the whole map (d), as predicted by point i). Under \ours{} the ego keeps its source statistics, and the measured change of its features averages $7.3\times10^{-5}$, about 220 times smaller than the 0.016 under Pooled BN and attributable to floating-point differences of batched computation.

\subsection{Agent-Wise Statistics}
\label{sec:agentwise}
\ours{} estimates the statistics of each collaborator from its own slice. For every BN layer of $\Enc$ and every collaborator $i\geq1$,
\begin{equation}
\hat\bmu_i=\frac{1}{|X_i|}\sum_{x\in X_i}x,\qquad
\hat\bsig^2_i=\frac{1}{|X_i|}\sum_{x\in X_i}(x-\hat\bmu_i)^2,
\label{eq:agentstats}
\end{equation}
where the sums run over the batch and spatial positions of the slice, separately for each channel. The collaborator's slice is then normalized as $\BN(x;\hat\bmu_i,\hat\bsig_i^2)$ with the original $\gamma$ and $\beta$. In the image backbone, $X_i$ holds the $C$ camera images of agent $i$, so the statistics are pooled over the views of one agent, which share the same corruption. In the BEV backbone, $X_i$ is the agent's BEV map and the statistics are taken over its spatial positions. Layers whose input cannot be split into per-agent slices keep the source statistics.

We use the statistics of the current frame rather than a running average over the test stream for two reasons. First, the set of collaborators and their conditions change from frame to frame, and a running average would carry statistics of one collaborator over to another. Second, frame-wise statistics make the output of a frame independent of the frames processed before it, so the method has no state that can drift. Section~\ref{sec:ablation} compares the two choices.

\subsection{Ego Anchoring}
\label{sec:anchor}
For the ego, \ours{} keeps the source statistics, $\BN(x;\bmu_\src,\bsig_\src^2)$. Under the setting of Section~\ref{sec:problem} the ego's input follows $P_\src$, so the source statistics are the correct ones, and the ego's own features are then exactly those of the pre-trained model. This guarantees that the adaptation of collaborators cannot alter the ego's own evidence, which is the failure mode observed for Tent in Section~\ref{sec:analysis}. When the ego itself may be corrupted, the ego can use its own statistics as well; we call this the \emph{symmetric} variant and evaluate it in Section~\ref{sec:ablation}.

\subsection{Properties and Deployment}
Algorithm~\ref{alg:agentnorm} summarizes the procedure. \ours{} has no learnable parameters and performs no backward pass, so it cannot diverge in the way observed for entropy minimization. It requires no labels, no source data and no information about the corruption. The per-agent mean and variance in \eqref{eq:agentstats} are the same quantities that a BN layer computes in training mode, so the overhead is a few reductions per layer and no additional memory. Because each slice $X_i$ belongs to one agent, the computation for collaborator $i$ can run on that agent's own processor before transmission: the collaborator simply switches its encoder's BN layers to its own statistics, the transmitted feature keeps its shape and size, and the ego needs no extra message. The same module wraps any intermediate-fusion model whose per-agent encoder uses BN, without changing its weights.

\begin{algorithm}[t]
\caption{\ours{} for one frame}
\label{alg:agentnorm}
\begin{algorithmic}[1]
\REQUIRE pre-trained encoder $\Enc$ with source statistics $(\bmu_\src,\bsig^2_\src)$ in every BN layer; images $\{\mathcal{I}_i\}_{i=0}^{N-1}$
\STATE run $\Enc$ on all agents, with every BN layer computed as follows:
\FOR{each BN layer of $\Enc$ with input $X=[X_0;\dots;X_{N-1}]$}
\STATE $\tilde X_0\leftarrow\BN(X_0;\bmu_\src,\bsig^2_\src)$ \hfill (ego anchoring)
\FOR{$i=1$ to $N-1$}
\STATE compute $(\hat\bmu_i,\hat\bsig^2_i)$ on $X_i$ by \eqref{eq:agentstats}
\STATE $\tilde X_i\leftarrow\BN(X_i;\hat\bmu_i,\hat\bsig^2_i)$ \hfill (agent-wise statistics)
\ENDFOR
\ENDFOR
\STATE collect $F_i$ from the encoder; $\hat Y\leftarrow\Head(\Fuse(F_0,\{\mathcal{T}_{i\to0}(F_i)\}))$
\ENSURE detections $\hat Y$; no parameter or statistic is stored
\end{algorithmic}
\end{algorithm}

\section{Experiments}
\label{sec:exp}

\subsection{Experimental Setup}
\label{sec:setup}

\textbf{Datasets and models.} Our main benchmark is the camera track of OPV2V~\cite{xu2021opv2v}, a simulated vehicle-to-vehicle dataset in which every vehicle carries four cameras. We use the full test split of 2,170 frames, 2,051 of which contain at least one collaborator. We evaluate four public camera checkpoints released with HEAL~\cite{lu2024heal}: AttFuse~\cite{xu2021opv2v}, V2X-ViT~\cite{xu2022v2xvit}, DiscoNet~\cite{li2021disconet}, and a checkpoint released under the F-Cooper name whose configuration uses attention fusion. We denote the last one AttFuse-B and treat it as a second, independently trained attention-fusion model rather than as evidence about max-pooling fusion. All four share a Lift-Splat-Shoot encoder with an EfficientNet-B0 image backbone, so their BN layers are all in the per-agent encoder. To test real-world data, we use the cooperative part of DAIR-V2X-C~\cite{yu2022dairv2x}, in which a vehicle cooperates with a roadside unit, each with one camera. No camera checkpoint is released for it, so we trained the camera-only CoAlign~\cite{lu2023coalign} configuration of HEAL for 30 epochs and kept the epoch with the best validation loss; evaluation uses the 1,789 validation frames.

\begin{figure*}[t]
\centering
\includegraphics[width=0.9\textwidth]{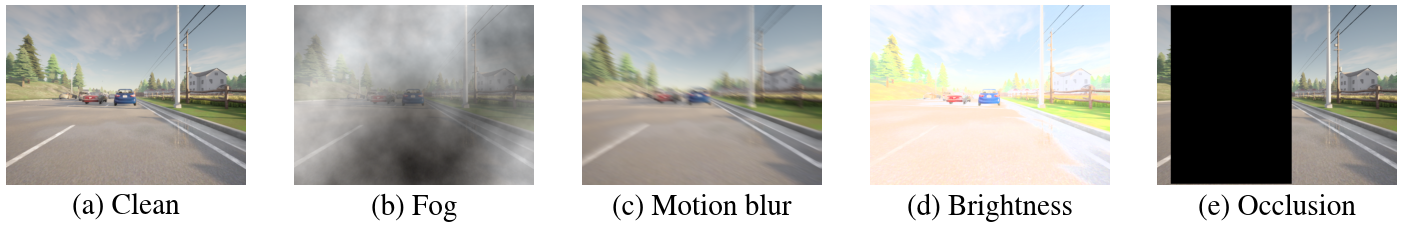}
\caption{One collaborator camera image of OPV2V under the four corruptions at severity 5. Only the collaborators' cameras are corrupted; the ego's cameras stay clean.}
\label{fig:corruptions}
\end{figure*}

\textbf{Corruptions.} Following RCP-Bench~\cite{du2025rcpbench}, we apply image corruptions from the common-corruption suite~\cite{hendrycks2019benchmarking} to the cameras of the collaborators only; the ego's cameras stay clean unless stated otherwise. On OPV2V we use fog, motion blur and brightness, and an occlusion corruption that pastes a black rectangle at a random position covering one quarter (severity 3) or one half (severity 5) of each image. On DAIR-V2X-C we corrupt the roadside camera with fog, snow, frost, brightness and motion blur. All corruptions are applied at severities 3 and 5, and the random seed of each image is fixed by the frame, agent, camera, corruption type and severity, so every method sees identical inputs. Fig.~\ref{fig:corruptions} shows one camera image under each corruption at severity 5.

\textbf{Metrics.} We report average precision at IoU thresholds 0.5 and 0.7 (\apf{} and \aps{}, in \%) computed with the HEAL evaluation code within $\pm$51.2~m of the ego on OPV2V and $\pm$102.4~m on DAIR-V2X-C. Differences in AP are given in points. Our evaluation of the clean AttFuse checkpoint gives 56.81 \apf{}, whereas the evaluation file distributed with the checkpoint reports 52.91; we therefore compare only numbers produced by our own pipeline and never mix them with published values.

\textbf{Compared methods.} \emph{No adaptation} runs the source model. \emph{Ego only} discards all collaborators. \emph{Pooled BN} uses test-batch statistics over all agents of a frame, the test-time baseline of RCP-Bench~\cite{du2025rcpbench}. \emph{Feature-statistics alignment} updates the collaborator's BN affine parameters online so that the per-channel mean and standard deviation of its lifted and pre-fusion BEV features match those of clean training frames, in the spirit of ActMAD~\cite{mirza2023actmad}. \emph{Tent-style entropy minimization}~\cite{wang2020tent} and \emph{region gating} are described in Section~\ref{sec:analysis}. As a reference that is not a valid test-time method, the \emph{oracle} trains the collaborator's BN affine parameters online with the detection loss on the ground-truth labels of each frame.

\textbf{Implementation details.} \ours{} and Pooled BN are inference-time changes of the BN layers and use the released weights unchanged. On OPV2V, the ego and the collaborators are encoded by two copies of the encoder with identical weights: the ego copy keeps the source statistics, and the collaborator copy normalizes the collaborators of a frame with their own statistics. When a frame has several collaborators, they are normalized as one group; since all collaborators of a frame receive the same corruption in our protocol, this matches the per-agent computation in that case. On DAIR-V2X-C, where there is a single collaborator, we use a generic module that replaces every BN layer of the model and splits its input by agent. No adaptation, Pooled BN and \ours{} are deterministic and independent of the frame order, so each is evaluated once. The gradient-based methods use seed 0 for the frame order. All experiments run on a single NVIDIA RTX 3090 GPU.

\subsection{Main Results}
\label{sec:main}

\begin{table}[t]
\centering
\caption{Collaborator-only corruption at severity 5 on OPV2V (camera). \apf{} and \aps{} in \%. $\Delta$: \ours{} minus Pooled BN in \apf{} points. Best per row in bold.}
\label{tab:main}
\setlength{\tabcolsep}{2.0pt}
\footnotesize
\begin{tabular}{llccccccc}
\toprule
& & \multicolumn{4}{c}{\apf{}} & \multicolumn{3}{c}{\aps{}} \\
\cmidrule(lr){3-6}\cmidrule(lr){7-9}
Model & Corruption & \makecell{No\\adapt.} & \makecell{Pooled\\BN} & Ours & $\Delta$ & \makecell{No\\adapt.} & \makecell{Pooled\\BN} & Ours \\
\midrule
\multirow{4}{*}{AttFuse} & Fog & 37.25 & 40.65 & \textbf{51.98} & +11.34 & 19.04 & 18.08 & \textbf{24.35} \\
 & Motion blur & 36.48 & 42.90 & \textbf{49.83} & +6.93 & 18.59 & 18.94 & \textbf{22.88} \\
 & Brightness & 42.37 & 53.87 & \textbf{55.59} & +1.72 & 21.46 & 23.47 & \textbf{25.70} \\
 & Occlusion & \textbf{39.92} & 35.17 & 35.63 & +0.46 & \textbf{19.87} & 14.83 & 17.95 \\
\midrule
\multirow{4}{*}{V2X-ViT} & Fog & 29.69 & 45.92 & \textbf{58.86} & +12.94 & 13.14 & 20.16 & \textbf{27.98} \\
 & Motion blur & 33.51 & 46.36 & \textbf{55.45} & +9.10 & 15.32 & 21.01 & \textbf{26.61} \\
 & Brightness & 53.75 & \textbf{63.75} & 62.93 & -0.82 & 25.43 & \textbf{30.01} & 29.90 \\
 & Occlusion & 32.86 & \textbf{37.85} & 37.51 & -0.34 & 15.25 & 15.21 & \textbf{18.22} \\
\midrule
\multirow{4}{*}{DiscoNet} & Fog & 33.33 & 43.73 & \textbf{51.57} & +7.85 & 15.11 & 20.80 & \textbf{23.63} \\
 & Motion blur & 31.56 & 42.89 & \textbf{48.42} & +5.53 & 14.84 & 19.47 & \textbf{21.18} \\
 & Brightness & 44.11 & \textbf{56.43} & 54.03 & -2.40 & 19.21 & \textbf{25.95} & 25.16 \\
 & Occlusion & 35.30 & \textbf{37.52} & 33.55 & -3.96 & 15.59 & \textbf{16.79} & 15.27 \\
\midrule
\multirow{4}{*}{AttFuse-B} & Fog & 29.42 & 30.79 & \textbf{43.32} & +12.53 & 14.98 & 10.64 & \textbf{19.09} \\
 & Motion blur & 31.31 & 32.36 & \textbf{42.80} & +10.45 & 15.13 & 10.93 & \textbf{18.07} \\
 & Brightness & 38.92 & 45.54 & \textbf{48.01} & +2.47 & 18.07 & 16.85 & \textbf{20.89} \\
 & Occlusion & \textbf{33.71} & 19.24 & 27.93 & +8.69 & \textbf{15.58} & 7.72 & 13.10 \\
\bottomrule
\end{tabular}
\end{table}

Table~\ref{tab:main} reports the four checkpoints under the four corruptions at severity 5. Three patterns emerge.

\emph{Fog and motion blur.} \ours{} is best on all eight combinations of checkpoint and corruption, by 5.53 to 12.94 \apf{} points over Pooled BN and by 11.50 to 29.17 points over no adaptation. The largest gain is on V2X-ViT under fog, where \ours{} raises \apf{} from 29.69 to 58.86 while Pooled BN reaches 45.92. On AttFuse-B, pooling barely helps \apf{} (30.79 versus 29.42 under fog) and lowers \aps{} from 14.98 to 10.64, whereas \ours{} reaches 43.32 and 19.09. This is the behavior predicted by \eqref{eq:residual}: the pooled statistics move the clean ego and under-correct the collaborator.

\emph{Brightness.} \ours{} improves over no adaptation by 9.09 to 13.22 \apf{} points, and \ours{} and Pooled BN are within 2.47 points of each other, each winning on two checkpoints. Brightness shifts the collaborator's feature statistics less than fog (Fig.~\ref{fig:stats}(a)), so the mixture in \eqref{eq:pmu}--\eqref{eq:pvar} is closer to the collaborator's own statistics.

\emph{Occlusion.} Neither method is reliable. \ours{} is below no adaptation on three of the four checkpoints, and Pooled BN on two. A black rectangle removes image content instead of changing its statistics, and normalization cannot restore missing content. We return to this case in Section~\ref{sec:discussion}.

Averaged over fog, motion blur and brightness, \ours{} reaches 52.47, 59.08, 51.34 and 44.71 \apf{} on AttFuse, V2X-ViT, DiscoNet and AttFuse-B, compared with 45.81, 52.01, 47.68 and 36.23 for Pooled BN and 38.70, 38.98, 36.33 and 33.22 without adaptation. On AttFuse under fog, the oracle that adapts the collaborator with ground-truth supervision reaches 57.13 \apf{}; \ours{} closes 74\% of the gap between no adaptation (37.25) and this oracle without using any label.

\textbf{Clean collaborators.} Applying \ours{} when nothing is corrupted does not reduce accuracy relative to the source model: \apf{} changes from 56.81 to 57.85 on AttFuse and from 66.46 to 68.50 on V2X-ViT (the latter source value is the run of Section~\ref{sec:ablation} in which the switching rule never triggers, which is identical to no adaptation). Pooled BN is higher on clean data (59.17 and 71.12), and also on DiscoNet and AttFuse-B (60.68 versus 57.17, and 51.00 versus 50.27). We discuss this in Section~\ref{sec:discussion}.

\subsection{Robustness to Corruption Severity}

\begin{figure*}[t]
\centering
\includegraphics[width=\textwidth]{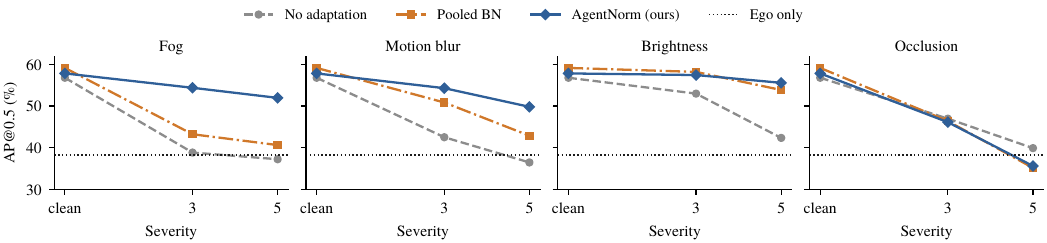}
\caption{\apf{} of AttFuse on OPV2V as the severity of the collaborator's corruption increases (ego clean). The dotted line is the ego-only model. Under fog and motion blur, \ours{} degrades much more slowly than Pooled BN and stays above the ego-only model; under occlusion, no method helps.}
\label{fig:severity}
\end{figure*}

Fig.~\ref{fig:severity} sweeps the severity on AttFuse. From clean collaborators to severity 5, \apf{} falls by 19.56 points under fog without adaptation and by 18.52 points with Pooled BN, but by 5.87 points with \ours{}. Under motion blur, the drops are 20.33, 16.27 and 8.02 points. Without adaptation, cooperation falls below the ego-only model at severity 5 for both corruptions, while \ours{} stays more than 11 points above it. Under brightness, \ours{} and Pooled BN remain close at both severities.

\subsection{Real-World Data}

\begin{table}[t]
\centering
\caption{Roadside-only corruption on DAIR-V2X-C (camera CoAlign, \apf{} in \%). For each severity: no adaptation / Pooled BN / \ours{}. Best per severity in bold. The vehicle-only model reaches 1.65.}
\label{tab:dair}
\setlength{\tabcolsep}{3.4pt}
\footnotesize
\begin{tabular}{lcccccc}
\toprule
& \multicolumn{3}{c}{Severity 3} & \multicolumn{3}{c}{Severity 5} \\
\cmidrule(lr){2-4}\cmidrule(lr){5-7}
Corruption & None & Pooled & Ours & None & Pooled & Ours \\
\midrule
Fog & 1.65 & 2.78 & \textbf{6.75} & 1.50 & 2.07 & \textbf{5.88} \\
Snow & 1.57 & 4.82 & \textbf{7.35} & 1.16 & 2.44 & \textbf{5.17} \\
Frost & 1.41 & 3.42 & \textbf{6.00} & 1.30 & 2.71 & \textbf{5.40} \\
Brightness & 5.61 & 7.25 & \textbf{8.82} & 1.17 & 3.79 & \textbf{6.08} \\
Motion blur & \textbf{8.97} & 7.55 & 8.93 & 6.86 & 6.01 & \textbf{7.46} \\
\midrule
Clean & \multicolumn{6}{c}{10.83 / 8.86 / 9.98 (no adapt. / pooled / ours)} \\
\bottomrule
\end{tabular}
\end{table}

Table~\ref{tab:dair} reports DAIR-V2X-C with the roadside camera corrupted. The absolute accuracy of camera-only detection on this dataset is low for the model we trained (10.83 \apf{} with clean inputs and 1.65 for the vehicle alone), so the results should be read as a check of direction rather than of magnitude. The direction agrees with OPV2V: \ours{} is above Pooled BN in all ten corrupted settings and above no adaptation in nine of them, the exception being motion blur at severity 3 (8.93 versus 8.97). Under fog, snow and frost at severity 5, roadside corruption pulls cooperative detection to 1.16--1.50 \apf{}, below the vehicle-only model, and \ours{} restores it to 5.17--5.88. With clean inputs, \ours{} is 0.85 points below the source model and 1.12 points above Pooled BN.

\subsection{Comparison with Alternative Remedies}

\begin{figure}[t]
\centering
\includegraphics[width=\linewidth]{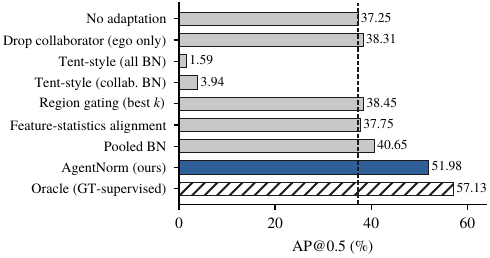}
\caption{Remedies for a corrupted collaborator (AttFuse, OPV2V, collaborator fog at severity 5). The dashed line marks no adaptation. The oracle uses ground-truth labels and is shown only as a reference.}
\label{fig:alternatives}
\end{figure}

Fig.~\ref{fig:alternatives} places \ours{} among the remedies discussed so far on the hardest AttFuse setting. Discarding the collaborator, region gating and feature-statistics alignment stay within 1.2 points of no adaptation, the Tent-style baseline collapses, and Pooled BN gains 3.40 points. \ours{} gains 14.73 points, and only the label-supervised oracle is higher. Feature-statistics alignment uses the same signal as \ours{}, the mismatch between test and source statistics, but tries to remove it by gradient steps on a shared affine transform; \ours{} removes it directly by renormalizing each agent.

\subsection{Qualitative Results}

\begin{figure*}[t]
\centering
\includegraphics[width=\textwidth]{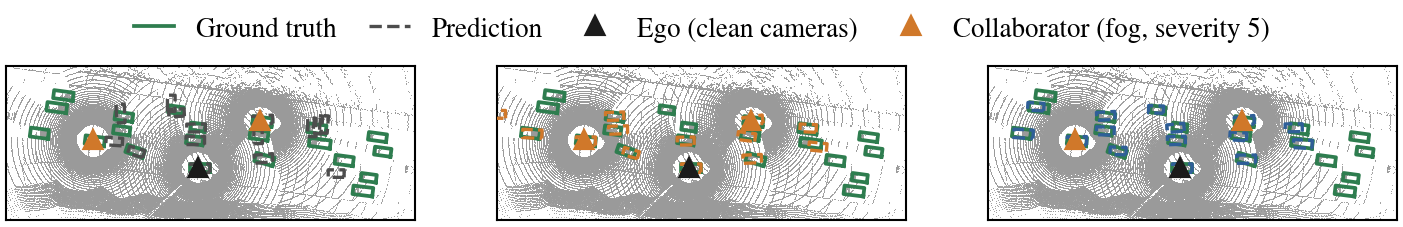}
\caption{Detections on an OPV2V test frame (AttFuse) with two collaborators (orange triangles) under severity-5 fog and a clean ego (black triangle). From left to right: no adaptation, Pooled BN and \ours{}. Solid green boxes: ground truth; dashed boxes: predictions. Gray points are the LiDAR scan provided by OPV2V, drawn only to show the road layout and the agents' surroundings; no model in this paper uses LiDAR. True positives at IoU 0.5: 6, 8 and 13 of 21 objects. The frame was selected as the one with the largest gain of \ours{} over no adaptation among 115 candidate frames (every 18th test frame); statistics over all candidates are given in the text.}
\label{fig:qual}
\end{figure*}

Fig.~\ref{fig:qual} shows the detections of the three methods on one test frame in which both collaborators are under severity-5 fog. Since a camera-only detector has no natural BEV background, we draw the LiDAR scan that OPV2V provides for the same frame as a map of the scene; it is not an input to any of the compared models. Without adaptation, the model detects 6 of the 21 objects at IoU 0.5, and several of its predictions are displaced from the ground truth. Pooled BN detects 8. \ours{} detects 13, and most of its remaining misses lie on the right side of the scene. This frame was chosen as the most favorable of 115 candidate frames (every 18th test frame), so it shows the best case rather than the typical one. Over all 115 candidates, the three methods detect 555, 621 and 747 of 1,390 objects at IoU 0.5 (39.9\%, 44.7\% and 53.7\%), and \ours{} detects more objects than no adaptation in 84 frames and fewer in 8.

\subsection{Ablation and Analysis}
\label{sec:ablation}

\begin{table*}[t]
\centering
\caption{Ablation on AttFuse (OPV2V). Only the collaborators are corrupted (severity 5) or nothing is corrupted (clean). Entries are \apf{} / \aps{} in \%. $^\dagger$Uses ground-truth labels; reference only.}
\label{tab:ablation}
\setlength{\tabcolsep}{5pt}
\footnotesize
\begin{tabular}{lccccc}
\toprule
Variant & Fog & Motion blur & Brightness & Occlusion & Clean \\
\midrule
No adaptation & 37.25 / 19.04 & 36.48 / 18.59 & 42.37 / 21.46 & 39.92 / 19.87 & 56.81 / 27.03 \\
Pooled BN & 40.65 / 18.08 & 42.90 / 18.94 & 53.87 / 23.47 & 35.17 / 14.83 & 59.17 / 27.25 \\
Agent-wise, symmetric & 53.11 / 23.55 & 50.69 / 22.24 & 56.60 / 25.20 & 37.06 / 17.48 & 58.76 / 26.69 \\
Agent-wise, ego-anchored (AgentNorm) & 51.98 / 24.35 & 49.83 / 22.88 & 55.59 / 25.70 & 35.63 / 17.95 & 57.85 / 27.18 \\
\midrule
AgentNorm, EMA statistics ($m{=}0.05$) & 51.12 / 24.17 & -- & -- & -- & -- \\
Oracle: GT-supervised collab.\ BN$^\dagger$ & 57.13 / 26.75 & -- & -- & -- & -- \\
\bottomrule
\end{tabular}
\end{table*}

\textbf{Is ego anchoring necessary?} Table~\ref{tab:ablation} compares \ours{} with the symmetric variant, in which the ego also uses its own statistics. The symmetric variant is 0.86 to 1.43 \apf{} points higher on all five columns, but \ours{} is 0.47 to 0.80 \aps{} points higher on all five. Both are far above Pooled BN under fog and motion blur. The choice is therefore a trade-off between recall at loose overlap and localization quality, not a large effect; we keep ego anchoring as the default because it leaves the ego's own features identical to the source model, which is the property that protects the ego from a failing collaborator (Section~\ref{sec:analysis}).

\textbf{Where should agent-wise statistics be applied?} The per-agent encoder has BN layers in the image backbone and in the BEV backbone. Fig.~\ref{fig:mechanism}(b) applies agent-wise statistics to only one of them. Applying them only in the image encoder gives 51.92 \apf{} under fog and 49.39 under motion blur, within 0.44 points of applying them everywhere (51.98 and 49.83). Applying them only in the BEV backbone gives 31.37 and 31.43, which is 5.88 and 5.05 points below no adaptation. The gain thus comes from the image encoder, where the corruption acts on image appearance; renormalizing only the BEV features, which are computed from images that were normalized with mismatched statistics, does not correct the shift and makes the features worse.

\textbf{Frame-wise or running statistics?} Replacing the frame-wise collaborator statistics by an exponential moving average over the test stream (momentum 0.05, in the style of DUA~\cite{mirza2022dua}) gives 51.12 \apf{} under fog, 0.86 points below the frame-wise version, and makes the result depend on the order of frames.

\textbf{Where does the gain over pooling come from?} Section~\ref{sec:pooled} predicts that per-agent statistics help only when the agents' distributions differ. Table~\ref{tab:both} tests this by corrupting the ego and the collaborators with the same corruption. Without adaptation, cooperation collapses (0.66 to 11.45 \apf{}). Agent-wise statistics (symmetric variant) and Pooled BN now perform within 1.01 \apf{} points of each other on all four corruptions. Together with Table~\ref{tab:main}, this shows that the advantage of \ours{} comes from the heterogeneity of the shift across agents.

\begin{table}[t]
\centering
\caption{Ego and collaborators corrupted with the same corruption (AttFuse, OPV2V, severity 5). \apf{} / \aps{} in \%.}
\label{tab:both}
\setlength{\tabcolsep}{3.5pt}
\footnotesize
\begin{tabular}{lccc}
\toprule
Corruption & No adapt. & Pooled BN & Agent-wise (sym.) \\
\midrule
Fog & 2.44 / 1.63 & 41.74 / 15.39 & 41.71 / 15.54 \\
Motion blur & 0.66 / 0.33 & 32.50 / 10.31 & 33.34 / 10.11 \\
Brightness & 11.45 / 5.26 & 47.98 / 20.12 & 46.97 / 19.16 \\
Occlusion & 2.31 / 0.83 & 6.25 / 3.00 & 6.07 / 3.11 \\
\bottomrule
\end{tabular}
\end{table}

\textbf{Should normalization be switched on only under shift?} Since \ours{} is applied to every collaborator, we also tested a rule that switches a collaborator to its own statistics only if the Gaussian KL divergence between its first-layer statistics and the source statistics exceeds the 99th percentile measured on clean validation frames. The rule is unreliable. Under motion blur it triggers for only 5\% (AttFuse) and 4.2\% (V2X-ViT) of collaborator frames, giving 36.99 and 34.13 \apf{} instead of 49.83 and 55.45 with \ours{}. Under fog on V2X-ViT it triggers for 98.6\% of frames and matches \ours{} (58.89 versus 58.86). Because \ours{} does not reduce accuracy on clean collaborators, applying it unconditionally is both simpler and more accurate than gating it with a shift detector.

\subsection{Efficiency}

\begin{table}[t]
\centering
\caption{Per-frame latency (ms) and peak GPU memory (MB) of AttFuse on OPV2V, collaborator fog at severity 5, including post-processing. 200 frames, the first 20 discarded, one RTX 3090.}
\label{tab:latency}
\setlength{\tabcolsep}{4pt}
\footnotesize
\begin{tabular}{lcccc}
\toprule
Method & Mean & Median & P95 & Memory \\
\midrule
No adaptation & 254.32 & 184.38 & 495.70 & 2669 \\
No adaptation (repeat) & 255.19 & 185.33 & 496.40 & 2669 \\
Pooled BN & 255.28 & 184.73 & 497.33 & 2669 \\
AgentNorm (ours) & 256.65 & 186.16 & 501.99 & 2669 \\
\bottomrule
\end{tabular}
\end{table}

Table~\ref{tab:latency} reports the runtime. \ours{} adds 2.33~ms to the mean latency of 254.32~ms (0.9\%), comparable to the 0.87~ms difference between two repeated runs of the unmodified model, and peak memory is unchanged. It adds no parameter, no gradient computation and no transmitted data. For comparison, the Tent-style baseline with all BN layers trainable required up to 20.2~GB of GPU memory in our runs, against 3.6~GB for the unadapted model in the same evaluation harness.

\section{Discussion and Limitations}
\label{sec:discussion}

\textbf{Content loss is not a statistics shift.} \ours{} corrects shifts in the first- and second-order statistics of a collaborator's features. Occlusion also shifts these statistics (Fig.~\ref{fig:stats}), but its main effect is that half of each image carries no information, and renormalizing the remaining features cannot recover it. On three of four checkpoints, \ours{} is below no adaptation under occlusion at severity 5. A deployment that expects lens blockage or heavy dirt should combine \ours{} with a mechanism that detects missing content and lowers the weight of the affected collaborator in fusion.

\textbf{Homogeneous shift.} When all agents undergo the same corruption, the per-agent and pooled statistics coincide in expectation, and Table~\ref{tab:both} confirms that \ours{} then brings no gain over Pooled BN. Its benefit is specific to shifts that differ between agents, which is the common case for roadside units and for vehicles that see different weather or lighting, but not the only one.

\textbf{Clean data.} On clean OPV2V data, Pooled BN is 0.73 to 3.51 \apf{} points above \ours{} on the four checkpoints, while \ours{} stays at or above the source model where we measured it. One possible explanation is a mild shift between the OPV2V training and test splits that affects all agents alike, for which pooled statistics, estimated from more samples, are the better estimator. We have not verified this explanation. In practice, the choice between the two estimators would require knowing whether the shift is shared, which is exactly the information that is unavailable at test time; under the shifts we consider, the loss of \ours{} on clean data is small compared with its gain under fog and motion blur.

\textbf{Scope of the evaluation.} We evaluated camera-based models only and make no claim about LiDAR encoders, whose BN statistics depend on point density and may respond differently to test-batch statistics. \ours{} applies to encoders with BN layers, which covers the camera models of the HEAL model zoo but not encoders that use only layer or group normalization; extending it to such encoders would require a different estimator of per-agent statistics. Our real-world experiment uses a camera-only DAIR-V2X-C model with low absolute accuracy, which limits how much its numbers can say about deployed systems. We evaluated synthetic corruptions with fixed severity for whole sequences, and the per-agent statistics rely on one frame of a few images per agent; very small inputs, such as a single low-resolution camera, would give noisier statistics. Finally, the gradient-based baselines were run with a single seed. Their collapse is far larger than seed variation, but their exact values should not be over-interpreted.

\section{Conclusion}
\label{sec:conclusion}

We studied camera-based cooperative perception when only some agents are corrupted. In this setting, a corrupted collaborator can pull cooperative detection below the ego-only model, region-level gating and a Tent-style entropy baseline do not help, and pooled test-time normalization mixes the statistics of clean and corrupted agents, lowering even the clean ego's own detection by up to 6.68 \apf{} points; this contamination arises in the image encoder. \ours{} normalizes each collaborator with the statistics of its own current observation and keeps the ego on its source statistics. It requires no training, gradients, labels or extra communication, adds about 2~ms per frame, and improves all four OPV2V checkpoints by large margins under fog and motion blur, with consistent gains over pooled normalization on real-world DAIR-V2X-C. It does not address occlusion-type content loss and offers no advantage when all agents share the same shift. Future work includes combining agent-wise normalization with fusion weights that account for missing content, extending it to encoders without BN, and validating it on cooperative data recorded in real adverse weather.


\end{document}